\documentclass[letterpaper, 10 pt, journal, twoside]{IEEEtran}  

\IEEEoverridecommandlockouts                              

\usepackage{graphicx}
\usepackage{makecell}
\usepackage{color}
\usepackage{cite}
\usepackage{hyperref}
\hypersetup{
hidelinks,
colorlinks=true,
linkcolor=red,
citecolor=green,
filecolor=magenta,
urlcolor=blue
}
\usepackage{multirow}
\usepackage{enumerate}
\usepackage{bm}
\usepackage{booktabs}
\usepackage{blindtext}
\usepackage[T1]{fontenc}
\usepackage{mathptmx} 
\DeclareMathAlphabet{\mathcal}{OMS}{cmsy}{m}{n}
\usepackage{times} 
\usepackage{amsmath} 
\usepackage{amssymb}  
\usepackage[ruled,linesnumbered]{algorithm2e}

\usepackage{xcolor}

\usepackage{bbm}
\usepackage{booktabs}
\usepackage{upgreek}
\usepackage{mathtools,xparse}
\usepackage{colortbl}

\newcommand{\PreserveBackslash}[1]{\let\temp=\\#1\let\\=\temp}
\newcolumntype{C}[1]{>{\PreserveBackslash\centering}p{#1}}
\newcolumntype{R}[1]{>{\PreserveBackslash\raggedleft}p{#1}}
\newcolumntype{L}[1]{>{\PreserveBackslash\raggedright}p{#1}}

\DeclarePairedDelimiter{\norm}{\lVert}{\rVert}
\NewDocumentCommand{\normL}{ s O{} m }{%
  \IfBooleanTF{#1}{\norm*{#3}}{\norm[#2]{#3}}_{L_2($\Omega$)}%
}

\DeclareFontFamily{U} {cmmi}{}

\DeclareFontShape{U}{cmmi}{m}{n}{
  <-6> cmmi5
  <6-7> cmmi6
  <7-8> cmmi7
  <8-9> cmmi8
  <9-10> cmmi9
  <10-12> cmmi10
  <12-> cmmi12}{}

\DeclareSymbolFont{Xcmmi} {U} {cmmi}{m}{n}

\DeclareMathSymbol{\psi}{\mathord}{Xcmmi}{32}

\makeatletter

\newcommand{\Rmnum}[1]{\expandafter\@slowromancap\romannumeral #1@}
\makeatother

\begin{document}
\title{Vision-Based Autonomous Car Racing Using Deep\\Imitative Reinforcement Learning}

\author{Peide~Cai,
        Hengli~Wang,
        Huaiyang~Huang,
        Yuxuan~Liu,
        and~Ming~Liu,~\IEEEmembership{Senior Member,~IEEE}
\thanks{This work was supported by Zhongshan Municipal Science and Technology Bureau Fund, under project ZSST21EG06, Collaborative Research Fund by Research Grants Council Hong Kong, under Project No. C4063-18G, and Department of Science and Technology of Guangdong Province Fund, under Project No. GDST20EG54, awarded to Prof. Ming Liu. \textit{(Corresponding author: Ming Liu.)}}
\thanks{The authors are with The Hong Kong University of Science and Technology, Hong Kong SAR, China (e-mail: pcaiaa@connect.ust.hk; hwangdf@connect.ust.hk; hhuangat@connect.ust.hk; yliuhb@connect.ust.hk; eelium@ust.hk). }
}


\maketitle

\begin{abstract}
Autonomous car racing is a challenging task in the robotic control area. Traditional modular methods require accurate mapping, localization and planning, which makes them computationally inefficient and sensitive to environmental changes. Recently, deep-learning-based end-to-end systems have shown promising results for autonomous driving/racing. However, they are commonly implemented by supervised imitation learning (IL), which suffers from the distribution mismatch problem, or by reinforcement learning (RL), which requires a huge amount of risky interaction data. In this work, we present a general deep imitative reinforcement learning approach (DIRL), which successfully achieves agile autonomous racing using visual inputs. The driving knowledge is acquired from both IL and model-based RL, where the agent can learn from human teachers as well as perform self-improvement by safely interacting with an offline world model. We validate our algorithm both in a high-fidelity driving simulation and on a real-world 1/20-scale RC-car with limited onboard computation. The evaluation results demonstrate that our method outperforms previous IL and RL methods in terms of sample efficiency and task performance. Demonstration videos are available at \url{https://caipeide.github.io/autorace-dirl/}.
\end{abstract}


\section{Introduction}

\IEEEPARstart{A}{utonomous} car racing is an attractive yet difficult robotic task. This task has two main challenges: First, the system has to control the car near its handling limits in highly nonlinear operating regimes\cite{liniger2015optimization}; and second, it has to avoid collisions in real time against dynamically and rapidly changing situations. Classical methods within this topic split the task into many building blocks, including mapping, localization, planning and control \cite{Kabzan2020AMZDT}. However, these methods are computationally inefficient, sensitive to subtle changes in the environment and prone to error propagation \cite{yurtsever2019survey}. Another approach to autonomous racing focuses on model predictive control (MPC) of racing cars and assumes the location information is known using expensive sensors such as accurate GPS, IMU and motion capture systems\cite{brunner2017repetitive, Liniger2017OptimizationBasedAR}. Because of these costly hardware requirements, the cars can only operate in rather controlled environments. 

Recent advances in deep learning have brought new approaches to handling autonomous racing. One popular method is called end-to-end driving\cite{bojarski2016end, codevilla2018end, cai2020probabilistic, pan2020imitation, Weiss2020DeepRacingPT, liu2020using, kendall2019learning, Cai2020VTGNetAV}. Based on the powerful representation abilities of deep neural networks, end-to-end methods can directly take as input raw high-dimensional sensory data (e.g., RGB images and Lidar pointclouds) and output low-level control commands (e.g., steering and throttle). Most of these are trained with imitation learning (IL), which can efficiently extract driving knowledge from expert demonstrations (i.e., observation-action data pairs), but suffers from the distribution mismatch problem (a.k.a \textit{covariate shift}). Although one can use DAgger\cite{ross2011reduction} or online supervision\cite{pan2020imitation} to alleviate such an issue, these two solutions bring further problems of costly labeling and intensive usage of privileged information (e.g., precise state of the ego-car). 

\begin{figure}[t]
        \centering
        \setlength{\abovecaptionskip}{0cm} 
        \includegraphics[width = \columnwidth]{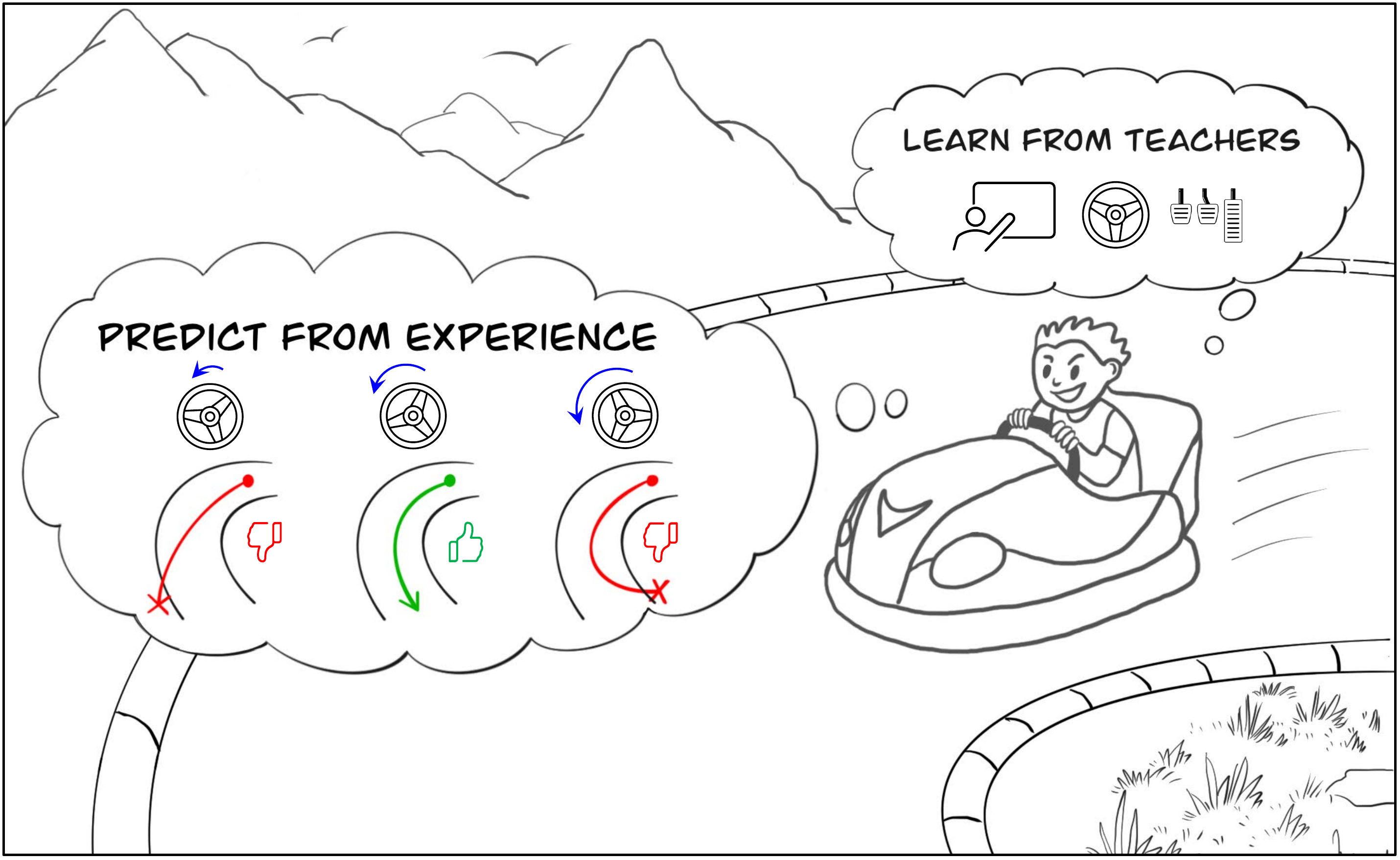}
        \caption{Humans drive based on prior knowledge (e.g., from teachers) and action-conditioned predictions of the world. We adopt this idea for an autonomous RC-car racing application by unifying IL and model-based RL.}
        \label{fig:cartoon}
        \vspace{-0.5cm}
\end{figure}

\begin{figure*}[t]
        \centering
        \setlength{\abovecaptionskip}{0cm}
        \includegraphics[width = 2\columnwidth]{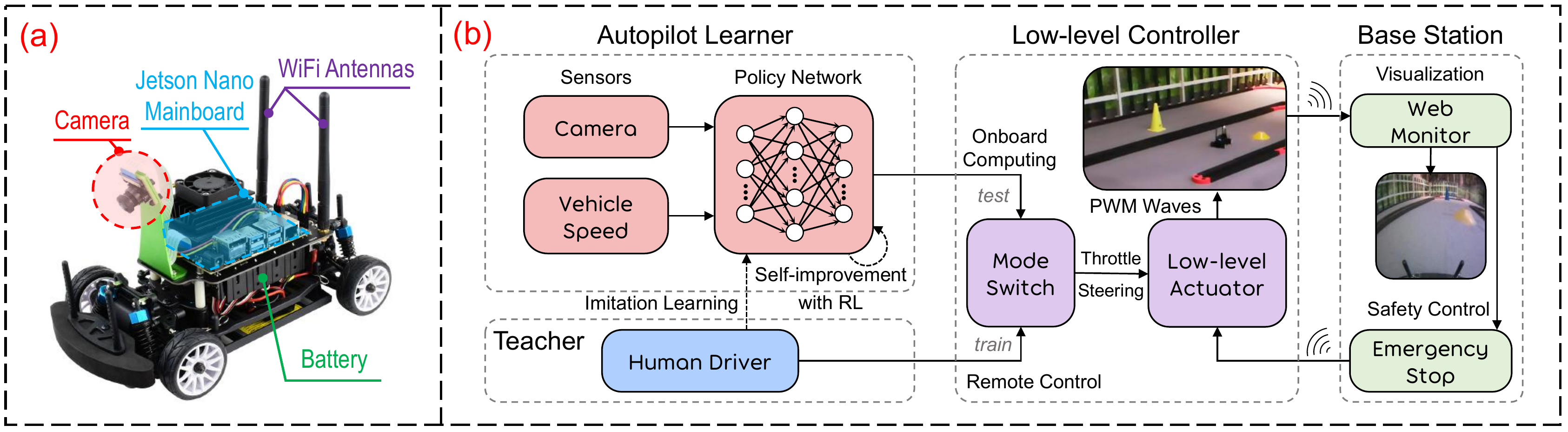}
        \caption{Overview of the (a) hardware and (b) software architecture of our RC-car racing system.}
        \label{fig:system}
        \vspace{-0.3cm}
\end{figure*}

Another learning-based approach with potential for application to autonomous car racing is RL. This paradigm drives an agent to \textit{interact} with the environment and gain knowledge from \textit{experience}, leading to better generalization performance through self-exploration, without relying on expert data. However, the two branches of RL, i.e., model-free and model-based RL, have their respective problems also. Model-free RL\cite{Cai2020HighSpeedAD, Zhu2019VisionbasedCI} is quite sample inefficient, which means the robot may fail millions of times to derive a good policy. Therefore, the training process is both expensive and dangerous for safety-critical tasks such as autonomous car racing. On the other hand, model-based RL \cite{deisenroth2011pilco, kamthe2018data} is commonly based on Gaussian progresses (GPs), which scale poorly on high-dimensional data such as RGB images. Due to these limitations, RL-based methods are mostly trained and tested in simulated, and, often, gaming environments, like Atari\cite{world_models, chua2018deep, Kaiser2020Model, hafner2019learning}, where physical robot damage is not a concern and ground-truth state information can be easily queried, leaving it unclear if these methods can scale to more challenging real-world tasks.

In this paper, we investigate learning-based methods by complementing IL and RL for the task of agile autonomous racing of a 1/20-scale RC-car. We first introduce \textit{Reveries-net}, which learns a spatio-temporal world representation from high-dimensional sensory inputs using uncertainty-aware recurrent neural networks (RNNs). Then, an IL-initialized end-to-end policy is reinforced using \textit{reveries}, i.e., imagined agent-world interactions, without the need of training in the real world. This paradigm, as depicted in Fig. \ref{fig:cartoon}, is similar to how humans drive, as we learn from both coaches and experienced \textit{what-ifs}: how the world might change if we take certain actions. The key contributions of this paper are summarized as follows:

\begin{enumerate}
    \item We propose an uncertainty-aware deep network, named \textit{Reveries-net}, to learn an action-conditioned probabilistic world model depicting how the future might unroll from accumulated experience.
    
    \item We combine IL and model-based RL into a unified learning framework, named deep imitative reinforcement learning (DIRL), to train end-to-end visual control policies.
    
    \item We validate the proposed method both in a realistic driving simulation and on a real-world 1/20-scale RC-car for autonomous racing, showing that it outperforms previous IL and RL baselines in terms of sample efficiency and task performance.
   
\end{enumerate}
\vspace{-1pt}
\section{Related Work}

\subsection{Imitation Learning}

Compared with classical modular systems for autonomous cars \cite{Kabzan2020AMZDT}, which are computationally expensive and sensitive to environmental changes, the advantage of IL is the speed of training and not needing to specify how the task should be performed (as the desired behaviors are embedded in expert demonstrations). Due to this efficiency and simplicity, IL has been the preferred approach to train end-to-end control policies\cite{bojarski2016end, codevilla2018end, cai2020probabilistic, pan2020imitation, Weiss2020DeepRacingPT, liu2020using}. For example, based on convolutional neural networks (CNNs), Bojarski \textit{et al.}\cite{bojarski2016end} developed a self-driving model \textit{PilotNet}, which uses camera images to compute steering commands. However, it was only tested on simple obstacle-free roads at low vehicle speeds, rather than in the racing environment that we consider. Following this work, \cite{Weiss2020DeepRacingPT} and \cite{liu2020using} used IL in autonomous car racing tasks, benefited by eye gaze\cite{liu2020using} and RNNs\cite{Weiss2020DeepRacingPT}. However, these were only evaluated in simulated environments like TORCS rather in the more complicated real world.

Although the efficiency of IL is appealing, using it only to train end-to-end policies has a disadvantage known as \textit{distribution mismatch}. Specifically, IL assumes the dataset is i.i.d. and trains networks to predict actions offline that do not affect future states. Actually, this is a common assumption in supervised learning and works well in areas like computer vision; however, driving a car is inherently a \textit{sequential} task, and the actions predicted by the network indeed affect the future. This phenomenon breaks the i.i.d. assumption and can lead to a distribution shift between training and testing. Thus, the policy network may make mistakes on unfamiliar state distributions after deployment. 

To alleviate the aforementioned problem, Ross \textit{et al.}\cite{ross2011reduction} proposed DAgger for online imitation learning, where a human expert is required to iteratively label the operation data generated by the trained policy with optimal commands in an open-loop way. The new data is then aggregated to the dataset to refine the policy. However, humans depend on feedback to act responsively; without actions affecting states in real-time, it is quite unnatural to label data frame-by-frame. To tackle this problem, subsequent works proposed using pre-designed experts rather than humans to provide online supervision. For example, Pan \textit{et al.}\cite{pan2020imitation} trained an end-to-end policy for autonomous RC-car racing using an MPC controller. However, in this approach, expensive sensors such as a GPS and IMU are still needed by the algorithmic expert. 
Compared to \cite{pan2020imitation}, our method reduces the hardware burden by using low-cost sensors (cameras and speed sensors) throughout the experiments. Moreover, we adopt RL to tackle the distribution mismatch problem using \textit{internal} self-exploration of agents instead of \textit{external} dense supervision from human/algorithmic experts.
\vspace{-5pt}

\subsection{Reinforcement Learning}
Different from IL, RL trains policies to maximize the sum of future rewards through interactions with the environment, where the network’s prediction affects future states. Therefore, it is a natural fit for sequential tasks. However, current model-free RL algorithms are quite expensive to train, often requiring millions of trial-and-error time steps\cite{Kaiser2020Model}, and have the risk of permanently damaging the physical agents by applying a suboptimal policy during training. Therefore, they are often limited in simulated environments for safety-critical tasks like autonomous racing\cite{Cai2020HighSpeedAD}.

On the other hand, model-based RL reduces the sample complexity. It first learns a predictive model of the world, and then uses that model to make decisions\cite{deisenroth2011pilco, kamthe2018data,world_models, chua2018deep, Kaiser2020Model, hafner2019learning, Baheri2020VisionBasedAD}. The types of models vary for different tasks. GPs are the most popular model choice, especially for low-dimensional problems such as \textit{cartpole} $\in \mathbb{R}^4$ \cite{deisenroth2011pilco, kamthe2018data}, but they struggle to represent complex dynamical systems. By contrast, DNNs \cite{world_models, chua2018deep, Kaiser2020Model, hafner2019learning, Baheri2020VisionBasedAD} scale well to high-dimensional states but they are mostly used as deterministic models\cite{Baheri2020VisionBasedAD, Kaiser2020Model}, and are consequently prone to suffer from overfitting in early learning stages. After the model is learned, it can be used to conduct either decision-time planning (DTP) or background planning (BP). For DTP, the learned model is used to find the best action sequence at \textit{runtime} situations during deployment, which requires an intensive sampling procedure\cite{chua2018deep,hafner2019learning}. Differently, BP learns how to act for \textit{any} situation by directly optimizing a parameterized policy\cite{deisenroth2011pilco, world_models, Kaiser2020Model}. It performs faster than DTP\cite{Kahneman11thinking} and is more suitable for tasks requiring high real-time performance. 

Our method uses DNNs to learn a high-dimensional ($\mathbb{R}^{27,648}$) visual world model. Since we aim to deploy our algorithm on physical robots instead of purely in simulations\cite{world_models, chua2018deep, Kaiser2020Model, hafner2019learning, Baheri2020VisionBasedAD}, we follow the idea of BP to train control policies for real-time performance. Another aspect that distinguishes our method from others is that we first introduce evidential learning\cite{amini2020deep, sensoy2018evidential} into RL to capture model uncertainties. Evidential learning is reported to have a four times faster inference speed and better performance than sampling-based methods to capture uncertainty in computer vision tasks\cite{amini2020deep}. Therefore, it is suitable for resource-constrained areas like robotics, and we show its benefits through an ablation study in Sec. \ref{subsec:experiment_real}.

\begin{figure}[t]
        \centering
        \setlength{\abovecaptionskip}{0cm}
        \includegraphics[width = \columnwidth]{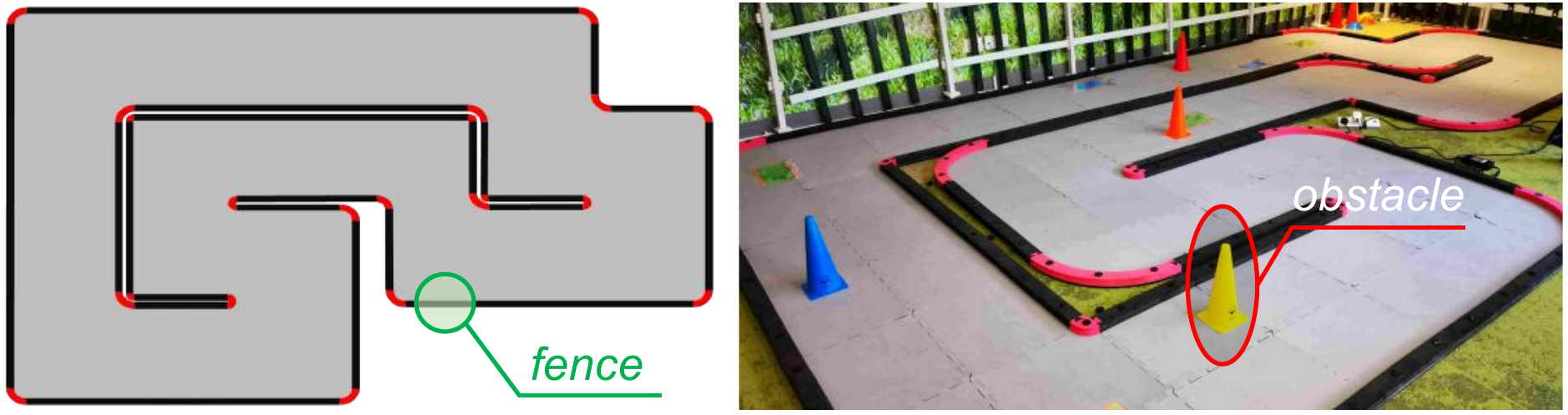}
        \caption{Track layout (6.7 m $\times$ 3.8 m). During experiments, multiple obstacles are placed on the track to construct challenging racing environments.}
        \label{fig:track}
        \vspace{-0.5cm}
\end{figure}

\subsection{Combining IL and RL in Robotics}

The complementarity between IL and RL has been motivating researchers to combine the benefits of both technologies. Current methods fall into two categories: 1) initializing the RL policy network with IL before starting exploration\cite{liang2018cirl}, and 2) loading the demonstration transitions into the replay buffer \cite{Zou2020AnEL, vecerik2017leveraging} to guide the RL process. Within these methods, the prior knowledge from supervised data provides a foundation for further self-optimization via RL, which can be regarded as \textit{the unity of knowledge and action}. However, previous methods have focused on the model-free RL techniques in this area. In this work we quantitatively show that learning an extra predictive model can not only reduce the training steps for convergence but also further improve the task performance (Sec. \ref{subsec:experiment_sim}).

\section{System Overview}
\label{sec:system}

We formulate the problem of racing car control as a high-agility autonomous driving task to be solved by an end-to-end policy network. Specifically, we consider a discrete-time partially observable Markov decision process (POMDP), where $\mathbb{S}$, $\mathbb{O}$ and $\mathbb{A}$ are the state, observation and action spaces, respectively. In our setting, the underlying state space is unknown, the observation $\bm{o}$ consists of camera images $\mathcal{I}$ and speed measurements $\bm{s}$, and action $\bm{a}$ consists of steering $\in[-1,1]$ and throttle $\in[0,1]$. The goal is to navigate the car around the given track (Fig. \ref{fig:track}) as fast as possible while avoiding collisions with the fence and random on-road obstacles. To this end, we develop a system (Fig. \ref{fig:system}) that can learn to perform fast and agile autonomous driving. The system is implemented on a 1/20-scale RC-car equipped with a low-cost embedded computer, Nvidia Jetson Nano. This computer has an ARM CPU @1.43 GHz, an integrated GPU and a 4 GB memory. In addition, a joystick can be used to remotely control the car manually, and on which a stop button is also configured to disable car motions in case of emergency. In our experiments, all computations run on the onboard Jetson Nano with a 5 W power mode in real-time (10 Hz).

\section{Methodology}
To solve the task of highly agile autonomous driving, the agent car should exploit prior knowledge as well as perform self-improvement based on accumulated experience. In our method, these two objectives are implemented by IL and model-based RL, respectively. In particular, we first train a world model depicting environment dynamics and a policy network from a human-gathered dataset. Then we iteratively refine these two networks in a model-based RL framework.

\begin{figure*}[t]
        \centering
        \setlength{\abovecaptionskip}{0cm}
        \includegraphics[width = 2\columnwidth]{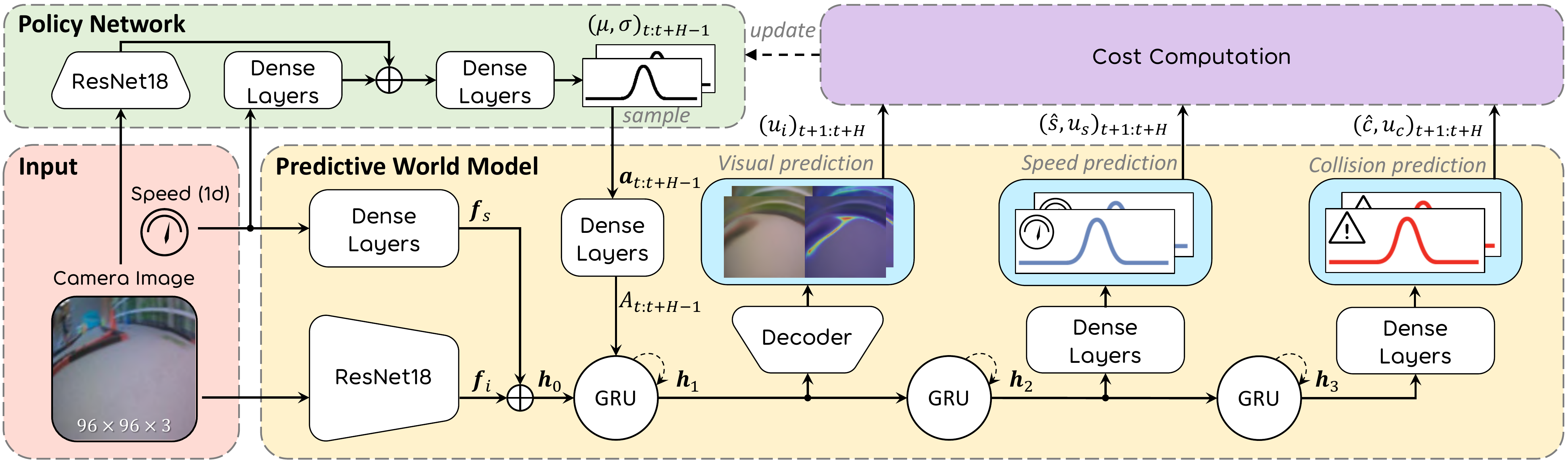}
        \caption{Network architecture for our end-to-end policy model and the action-conditioned predictive world model, \textit{Reveries-net}. Note that the large recurrent world model is used to provide imagined agent-world interactions (\textit{reveries}) with gradients to update the policies, and only the lightweight green shadowed area (policy) will be executed at test time. See Algorithm \ref{alg:dirl} for the pseudo code of the agent.}
        \label{fig:world-model}
        \vspace{-0.3cm}
\end{figure*}

\subsection{Data Collection}
\label{subsec:data_collection}

Based on the RC-car platform introduced in Sec. \ref{sec:system}, we first collect an expert-driving dataset, which is later used to initialize the world model (Sec. \ref{subsec:world-model}) and the policy network (Sec. \ref{subsec:policy-model}). Specifically, a human driver is asked to remotely control the steering and throttle of the RC-car at 10 Hz to make it run fast and safely on the track. For better training performance, we increase the diversity of the dataset in the following two ways: 1) we place obstacles at random locations on the track with random colors and numbers, and 2) following \cite{codevilla2018end}, we inject random noise $\in$ [-0.3, 0.3] on the throttle and steering for half of the data collection time, where the human has to recover the car from off-center or off-orientation errors in a timely manner. Note that due to these added difficulties and the requirement of fast car driving, even human experts make mistakes occasionally, causing the car to collide with the fence or obstacles. However, different from previous vanilla IL methods \cite{cai2020probabilistic, codevilla2018end} that completely ignore these failure cases, we believe that these driver control errors and related observations can provide indirect training signals within the model-based RL framework. Therefore, we still incorporate the failure cases into the dataset. Specifically, when a collision occurs, the binary collision signals $\bm{c}$ for the past 0.5 s are set to 1 and others to 0. Then the car will be reset for the next driving episode.

During driving, the agent records the camera RGB images $\mathcal{I}$ ($96\times96\times3$), car speeds $\bm{s}$, expert actions $\bm{a}^\star$, final actions $\bm{a}$ (may be masked with random noise over $\bm{a}^\star$), and collision signals $\bm{c}$. We finally collect 395 expert driving episodes, which last 2.6 hours in total with 94K frames.

\subsection{Predictive World Model}
\label{subsec:world-model}
\subsubsection{Model Architecture}
Based on the collected dataset, we first train an action-conditioned predictive world model \textit{Reveries-net} for later policy improvement. \textit{Reveries-net} takes as input the current observations (image $\mathcal{I}_t$, car speed $\bm{s}_t$) and a sequence of $H$ future actions $\bm{a}_{t:t+H-1}$. Then, it predicts states from $t+1$ to $t+H$, which are images $\Hat{\mathcal{I}}$, car speeds $\Hat{\bm{s}}$ and collision signals $\Hat{\bm{c}}$, along with their respective uncertainties $u_{i}$, $u_{{s}}$, and $u_{{c}}$.

The network details are shown in Fig. \ref{fig:world-model}, where a gated recurrent unit (GRU) architecture is used to predict forward in a latent space. The model first extracts environmental features $f_i\in\mathbb{R}^{512}$ and $f_s\in\mathbb{R}^{128}$ from observations $\mathcal{I}_t$ and $\bm{s}_t$, using the CNN backbone of ResNet18 and dense layers. Then, these features are concatenated to be the initial hidden state $h_0\in\mathbb{R}^{640}$ of the GRU. For sufficient expressive power, the model also processes the input actions with dense layers to produce higher dimensional action features $\bm{A}_{t:t+H-1}\in\mathbb{R}^{64 \times H}$, which are then sequentially fed into the three-layer GRU to generate hidden states $h_1$, $h_2$ and $h_3$ $\in \mathbb{R}^{512 \times H}$. For final predictions, a decoder with a combination of upsampling and convolutions is used to process $h_1$ for image reconstruction of $\Hat{\mathcal{I}}_{t+1:t+H}$, and dense layers are adopted to process $h_2$ and $h_3$ to predict $\Hat{\bm{s}}_{t+1:t+H}$ and $\Hat{\bm{c}}_{t+1:t+H}$, respectively.

\subsubsection{Uncertainty Estimation}
\label{subsubsec:world_model_train}
Different from previous deterministic models\cite{Baheri2020VisionBasedAD, Kaiser2020Model}, \textit{Reveries-net} explicitly considers the model uncertainty, to be better used for planning with the recent evidential deep learning\cite{amini2020deep, sensoy2018evidential}. We formulates learning as an evidence acquisition process and every training sample adds support to a learned higher-order, \textit{evidential} distribution. More specifically, for the regression tasks such as speed prediction of $\hat{\bm{s}}$, we follow the idea of \cite{amini2020deep} and place a higher-order normal inverse-gamma (NIG) distribution with hyperparameters $(\gamma_s, v_s, \alpha_s, \beta_s)$ over the likelihood functions. Then, the network is trained to infer these hyperparameters by jointly maximizing the model fit ($\mathcal{L}^\mathrm{NLL}_s$) and minimizing evidence of errors ($\mathcal{L}^\mathrm{R}_s$):
\begin{equation}
\begin{aligned}
\mathcal{L}^{\mathrm{NLL}}_s&=\frac{1}{2} \log \left(\frac{\pi}{v_s}\right)-\alpha_{s} \log \left(\Omega_{s}\right) +\log \left(\frac{\Gamma\left(\alpha_{s}\right)}{\Gamma\left(\alpha_{s}+1 / 2\right)}\right) \\
&+\left(\alpha_{s}+1/2\right) \log \left(\left(\bm{s}-\gamma_{s}\right)^{2} v_{s}+\Omega_{s}\right), \\
\mathcal{L}^{\mathrm{R}}_s&=\left|\bm{s}-\gamma_{s}\right| \cdot\left(2 \alpha_{s}+v_{s}\right), \quad \mathcal{L}_s = \mathcal{L}_s^{\mathrm{NLL}} + \lambda \mathcal{L}_s^{\mathrm{R}},
\end{aligned}
\end{equation}
where $\Gamma(\cdot)$ is the gamma function, $\bm{s}$ is the ground-truth future speed, $\lambda$ is the weight coefficient, and $\Omega_s = 2\beta_s(1+v_s)$. After model training, $\gamma_s$ will serve as the prediction of $\hat{\bm{s}}$, and the uncertainty $u_s$ can be calculated from Eq. (\ref{eq:uncertainty}), which is the sum of aleatoric (or data) uncertainty and epistemic (or model) uncertainty. This is the same for the visual prediction of $\hat{\mathcal{I}}$, leading to training loss $\mathcal{L}_i$ and image uncertainty $u_i$.
\begin{equation}
    u_s = \left(1+\frac{1}{v_s}\right)\frac{\beta_s}{(\alpha_s-1)}.
    \label{eq:uncertainty}
\end{equation}
For the classification task of predicting future collision signals $\hat{\bm{c}}$, we follow \cite{sensoy2018evidential} and replace the commonly used softmax layer with ReLU activation to output non-negative \textit{evidential} outputs $e_k$ for $k=1, \ldots, K$, where $K$ is the number of classes ($K=2$ in this work). Let ${y}$ be the one-hot vector encoding the ground-truth class of collision signal $\bm{c}$, with ${y}_{j}=1$ and ${y}_{k}=0$ for all $k \neq j$. Then, the training loss of collision prediction is defined as
\begin{equation}
\label{eq:lc}
    \mathcal{L}_c = \sum\nolimits_{j=1}^{K} {\left(y_{j}-\frac{e_{j}+1}{S} \right)^{2}}+\frac{(e_{j}+1)\left(S-e_j-1\right)}{S^{2}\left(S+1\right)},
\end{equation}
where $S=\sum\nolimits_{i=1}^{K}\left(e_{i}+1\right)$. Then, the class probability $\hat{p}_k$ and the overall uncertainty $u_c$ can be computed by
\begin{equation}
\label{eq:pk}
\hat{p}_{k}=({e_{k}+1})/{S},\ u_c={K}/{S}.
\end{equation}
For more details of the evidential deep learning, we refer readers to \cite{amini2020deep} and \cite{sensoy2018evidential}. Finally, the world model is trained end-to-end with the total loss $\mathcal{L}_w$, which is the weighted sum of $\mathcal{L}_i$, $\mathcal{L}_s$ and $\mathcal{L}_c$.

\begin{algorithm}[t]

\caption{DIRL\label{alg:dirl}}

Randomly initialize parameters of the policy $\pi_{\phi}$ and the world model $p_{\theta}$\;
Initialize dataset $\mathcal{D}$ by human experts (Sec. \ref{subsec:data_collection})\;
\tcp{Imitation learning}
Train policy $\pi_{\phi}$ using IL based on the collision-free samples and actions $\bm{a}^{\star}$ in $\mathcal{D}$ (Sec. \ref{subsubsec:policy_train})\;
\While{task not learned}{
\tcp{World model learning}
\While{not converge}{
    Draw $N$ data sequences $\{ (\bm{o}_t, {\bm{a}}_t, \bm{c}_t)_{t=k}^{k+H}\} \sim \mathcal{D}$\;
    Compute world model loss $\mathcal{L}_w$ (Sec. \ref{subsubsec:world_model_train})\;
    Update $\theta \leftarrow \theta - \alpha \nabla_{\theta} \mathcal{L}_w $\;
}
\tcp{Model-based policy refinement}
\While{not converge}{
    Draw $N$ data samples $\{\bm{o}_t\} \sim \mathcal{D}$\;
    Sample actions $\bm{a}_{t:t+H-1} \leftarrow \pi_{\phi}(\bm{o}_t)$ via Eq. \ref{eq:reparam}\;
    Predict future states with uncertainties $\hat{\bm{s}}, \hat{\bm{c}}, u_{(i,s,c)} \leftarrow p_{\theta}\left(\bm{o}_t, \bm{a}_{t:t+H-1} \right)$\;
    Compute policy loss $\mathcal{L}_{\pi}$ via Eq. \ref{eq:policy_loss}\; 
    Update $\phi \leftarrow \phi - \alpha \nabla_{\phi} \mathcal{L}_{\pi} $\;
}
\tcp{Data collection}
\For{time step $\leftarrow$ 1 \KwTo{T}}{
    Compute $\bm{a}_t \leftarrow \pi_{\phi}(\bm{o}_t)$ (Sec. \ref{subsubsec:policy_model_architecture})\;
    Collision $\bm{c}_t \leftarrow 1$ if emergency stop, else 0\;
}
$\mathcal{D}\leftarrow\mathcal{D}\cup\{\left(\bm{o}_{t}, \bm{a}_{t}, \bm{c}_{t}\right)_{t=1}^{T}\}$
}
\end{algorithm}

\subsection{End-to-End Policy Model}
\label{subsec:policy-model}
\subsubsection{Model Architecture}
\label{subsubsec:policy_model_architecture}
As shown in Fig. \ref{fig:world-model}, the policy $\pi$: $\mathbb{O}\mapsto\mathbb{A}$ processes observations $\bm{o}_t = \langle\mathcal{I}_t, \bm{s}_t\rangle$ using ResNet18 and dense layers, and outputs a Gaussian distribution with mean $\mu_{t: t+H-1}$ and variance $\sigma_{t: t+H-1}$. Then, a sequence of $H$ actions $\bm{a}_{t: t+H-1}$ can be sampled with the reparametrization trick for exploration in the learned world model:
\begin{equation}
\bm{a}_{t+h}=\mu_{t+h} + \sigma_{t+h} \cdot \epsilon, \quad \epsilon \sim \operatorname{Normal}(0, 1).
\label{eq:reparam}
\end{equation}
Note that during testing, only the mean of the first action, $\mu_t$, will be executed, which is similar to the MPC paradigm.

\subsubsection{Model Training}
\label{subsubsec:policy_train}
We first train the policy $\pi$ \textit{implicitly} based on IL with the L1 loss function, using the collision-free frames and expert actions $\bm{a}^\star$ in the collected dataset (Sec. \ref{subsec:data_collection}). Then, we refine the policy \textit{explicitly} by minimizing a loss function $\mathcal{L}_{\pi}$ over a finite horizon of length $H$:
\begin{equation}
\begin{aligned}
&\min _{\pi} \mathcal{L}_{\pi}, \quad \mathcal{L}_{\pi}=\mathbb{E}_{\rho_{\pi}}\left[\sum\nolimits_{t=1}^{H} J_t\right],
\label{eq:policy_loss}
\end{aligned}
\end{equation}
where $\rho_{\pi}$ is the distribution of the trajectory ($\bm{o}_1, \bm{a}_1,..., \bm{o}_H, \bm{a}_H$) generated by running policy $\pi$ in the world model, and $J_t$ is the weighted sum of the following three sub-costs to encourages fast and safe autonomous driving:

\textbf{Speed cost}: $J^{spd}_t = -\mathbb{E}_{\rho_{\pi}}\left[\sum\nolimits_{t=1}^{H} \hat{\bm{s}}_t\right]$, which is the negative expectation of future speeds to encourage fast driving.

\textbf{Collision cost}: $J^{coll}_t = \mathbb{E}_{\rho_{\pi}}\left[\sum\nolimits_{t=1}^{H} \hat{\bm{c}}_t\right]$, which is the expectation of predicted collision signals under policy $\pi$ to punish dangerous driving behaviors.

\textbf{Uncertainty cost}: $J^{unc}_t = \mathbb{E}_{\rho_{\pi}}\left[\sum\nolimits_{t=1}^{H} u_{i(t)} + u_{s(t)} + u_{c(t)}\right]$, which is the expectation of uncertainties from image, speed and collision predictions. We penalize these uncertainties to avoid overestimation of the learned world model.

Considering the agent may not initially visit all parts of the environment, we need to iteratively collect new experiences and refine the world and policy models. The final algorithm is shown in Algorithm \ref{alg:dirl}. 
\begin{figure}[t]
        \centering
        \setlength{\abovecaptionskip}{0cm}
        \includegraphics[width = \columnwidth]{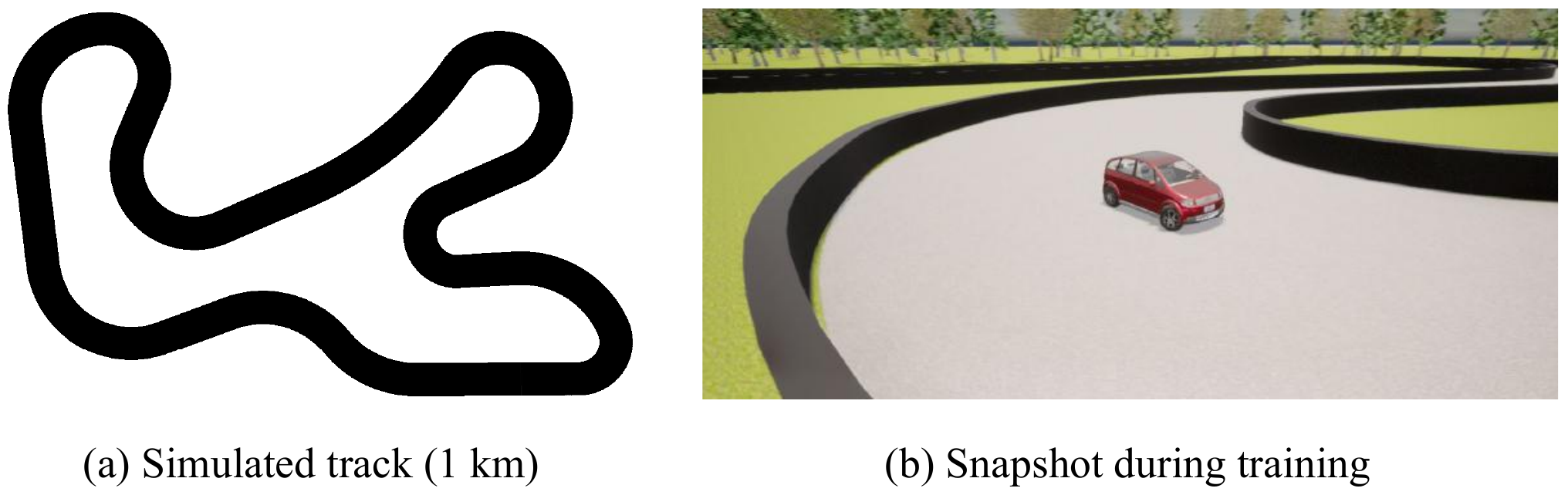}
        \caption{Overview of the simulated benchmarking environment.}
        \label{fig:sim}
        \vspace{-0.3cm} 
\end{figure}

\section{Experiments and Discussion}

In this section, we first evaluate the performance of our method by comparing it with existing approaches in a driving simulator (Sec. \ref{subsec:experiment_sim}). Then, we demonstrate a solution on the real-world RC-car and perform an ablation study to understand the importance of uncertainty estimation in our method (Sec. \ref{subsec:experiment_real}). For both simulation and real-world experiments we set the prediction horizon $H = 10$ for balancing the model size and prediction capability (equal to 1 s), and use the Adam optimizer for network training. For the world model, we use a learning rate $\alpha = 5e^{-4}$ and batch size $N=32$. For the policy network, we set $\alpha = 1e^{-3}$, $N=128$ during imitation learning, and $\alpha = 1e^{-5}$, $N=36$ during policy refinement. All methods run at 10 Hz.

\subsection{Comparative Study in Simulation}
\label{subsec:experiment_sim}

Considering that training and testing models in simulation enables fair systematic benchmarking, we conduct a comparative study in a 3D driving simulator before conducting experiments in the real world. Specifically, we first build a track (Fig. \ref{fig:sim}) using Unreal Engine 4 and CARLA\cite{dosovitskiy2017carla}, which is a high-fidelity open-source simulator for autonomous driving research. The data-collection procedure and sensor setup are similar to those reported in Sec. \ref{subsec:data_collection}, but we neglect the obstacles here. We choose the following methods for comparison:

\begin{itemize}
    \item \textbf{IL}: We train IL to mimic the expert's future 10 actions based on the collision-free expert data\cite{codevilla2018end}.
    \item \textbf{DDPG}: Deep deterministic policy gradient is a popular continuous, model-free RL method, which has recently been used successfully for visual lane-following tasks on a low-speed real car\cite{kendall2019learning}.
    \item \textbf{DDPGfD}: This is a method that combines IL and RL, which modifies the original DDPG by preloading some demonstration transitions into the replay buffer and keeping them forever when training DDPG\cite{vecerik2017leveraging, Zou2020AnEL}.
    \item \textbf{CIRL}: This is another method that combines IL and RL. It initializes the policy network using IL and conducts further model-free finetuning using DDPG \cite{liang2018cirl}.
\end{itemize}

For fair comparison, all these methods use the same observation-action space and backbone as ours does. To stimulate fast and safe driving, the reward for training RL-based agents is the sum of the vehicle speed, negative heading angle error and negative cross track error. The reward is -1000 as a penalty if the agent collides with the fence. Finally, we test the different methods five times on the track and calculate their average speed, top speed, completion ratio (the proportion of the safe driving distance before a collision event to the track length), and lap time.

\subsubsection{Quantitative Analysis}
The comparative results are shown in Table \ref{tab:sim_eval}. Our method finally takes 163K of data to converge. Accordingly, we collect the same amount of data to train IL, and report the performance of other RL-based methods from two aspects: 1) how they perform using the same amount of data as ours, and 2) how much data they need for convergence and what their final performance is.

\textbf{Sample efficiency: }DDPG performs worst on this metric, as it requires 1369K of data to converge. By combining IL and RL, DDPGfD and CIRL require less data (300$\sim$370K). Our method takes these methods one step further by incorporating a world model to help with policy learning, and requires 1.8$\sim$8.4 times less data than the other RL-based methods. \textbf{Task performance: }Our method achieves better performance than others on most metrics. By comparison among the models using the same amount of data, DIRL is the only model that can reach 100\% on completion ratio with the highest average speed. IL achieves a slightly higher top speed than ours but it can only safely drive 66.5\% of the lap. After being trained with more data to converge, CIRL improves the completion ratio of IL to 77.1\%. On the other hand, DDPG and DDPGfD achieve a 100\% completion ratio at the cost of lower scores on average and top speed. Therefore, they require a much longer lap time (88.4$\sim$134.7 s) than ours (50.1 s). 

The results show that our method is not only more data efficient than previous methods, but also beneficial for improving the final task performance. We accredit this improvement to the usage of the world model in an IL-and-RL framework to help reduce the distribution mismatch and explore the environment more efficiently.

\subsubsection{Robustness to Non-Expert Data}
In this section, we further examine the robustness of our method to non-expert demonstration data. In particular, we add two levels of Gaussian noise $\mathcal{N}(0, \sigma)$ to the original dataset with different $\sigma$ (0.5, 1) and train policy models using IL. After this, we use the learned world model to refine the policies in our framework to distill the final policy. The evaluation results are shown in Fig. \ref{fig:suboptim}. It can be seen that since IL purely mimics the provided data, it is rather sensitive to the demonstration level. For example, the completion ratio decreases from 66.5\% to 13.1\% as more noise is added. By contrast, our method can make up for the sub-optimal training data to a certain extent. Specifically, DIRL can maintain the completion ratio at 100\% at different demonstration levels, and maintains an average speed of 55.3 km/h at 1.0-Noi (36.8 km/h for IL).

\begin{table}[t]
\newcommand{\tabincell}[2]{\begin{tabular}{@{}#1@{}}#2\end{tabular}}
\newcommand{\NA}{---}
        \setlength{\abovecaptionskip}{-1pt}
        \definecolor{minigray}{rgb}{0.92, 0.92, 0.92}
        \caption{Comparative Evaluation Results. $\uparrow$ Means Larger Numbers Are Better, $\downarrow$ Means Smaller Numbers Are Better. The Bold Font Highlights the Best Results in Each Column.}
        \label{tab:sim_eval}
        \centering
        \begin{tabular}{l|l|llll}
        \hline
        \multirow{2}{*}{Model} & \multirow{2}{*}{Data} &\makecell[l]{Avg.\\Speed} &\makecell[l]{Top\\Speed} & \makecell[l]{Complet.\\Ratio } & \makecell[l]{Lap\\Time} \\
        & & $(km/h)$ $\uparrow$ &  $(km/h)$ $\uparrow$ & $(\%)$ $\uparrow$ &  $(s)$ $\downarrow$\\
        \hline
        \hline
        \multirow{2}{*}{DDPG\cite{kendall2019learning}} & 163K  &19.9 &30.1& 3.3 & --- \\
        & 1369K  & 40.7 & 45.1& \textbf{100.0} & 88.4\\
        \hline
        \multirow{2}{*}{DDPGfD\cite{Zou2020AnEL}} & 163K& 19.6 & 36.3 & 3.5  & --- \\
        & 370K  & 27.2 & 35.1& \textbf{100.0} & 134.7\\
        \hline
        \multirow{2}{*}{CIRL\cite{liang2018cirl}} & 163K  & 38.0  & 68.7& 7.8 & ---\\
        & 300K & 71.6 & 101.9& 77.1  & ---\\
        \hline
        IL\cite{codevilla2018end} & 163K  & 73.1 & \textbf{107.0}& 66.5 & ---\\
        \hline
        DIRL(\textit{ours})& 163K  & \textbf{73.7} & 105.8& \textbf{100.0} & \textbf{50.1}\\
        
        \hline
        \end{tabular}
\end{table}

\begin{figure}[t]
        \centering
        \setlength{\abovecaptionskip}{0cm}
        \includegraphics[width = \columnwidth]{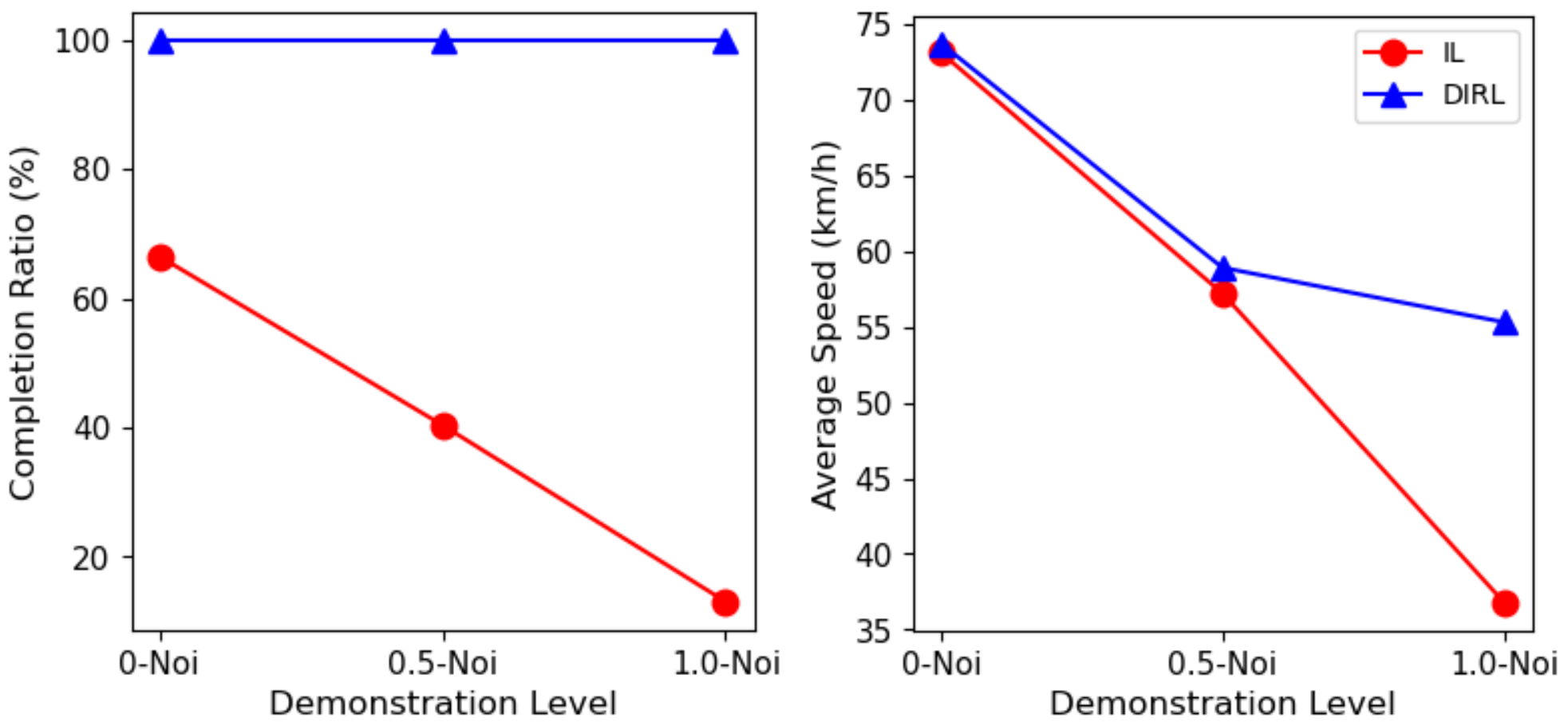}
        \caption{Robustness of our method to non-expert demonstration data. $\sigma$-Noi means we add extra Gaussian noise $\mathcal{N}(0,\sigma^2)$ to the original expert data.}
        \label{fig:suboptim}
        \vspace{-0.3cm}
\end{figure}

\subsection{Real-world Experiments}
\label{subsec:experiment_real}

\begin{table*}[t]
\newcommand{\tabincell}[2]{\begin{tabular}{@{}#1@{}}#2\end{tabular}}
\newcommand{\NA}{---}
        \setlength{\abovecaptionskip}{-1pt}
        \definecolor{minigray}{rgb}{0.92, 0.92, 0.92}
        \caption{Real-world Evaluation Results. $\uparrow$ Means Larger Numbers Are Better, $\downarrow$ Means Smaller Numbers Are Better. The Bold Font Highlights the Best Results in Each Column.}
        \label{tab:real_eval}
        \centering
        \begin{tabular}{l|l|cccc|cccc}
        \hline
        \multirow{2}{*}{Model} & \multirow{2}{*}{Data} & \multicolumn{4}{c|}{Easy (2 obstacles)}                         & \multicolumn{4}{c}{Hard (8 obstacles)}                         \\
        
        & & Avg. Speed$\uparrow$ & Top Speed$\uparrow$ & Interventions$\downarrow$ & Time$\downarrow$ & Avg. Speed$\uparrow$ & Top Speed$\uparrow$ & Interventions$\downarrow$ & Time$\downarrow$ \\
        \hline
        \hline
        
        DIRL(1 iter) & 92K & 0.81 m/s & 1.65 m/s & 4.67 & 82.93 s & \textbf{0.82 m/s} & \textbf{2.00 m/s} & 7.67 & 90.70 s \\
        
        DIRL(2 iter) & 104K & \textbf{0.82 m/s} & 1.76 m/s & 1.67 & 83.75 s & 0.76 m/s & 1.80 m/s & 5.00 & 89.71 s \\
        
        DIRL(3 iter) & 120K & 0.80 m/s & 1.66 m/s & \textbf{1.33} & \textbf{82.76 s} & 0.81 m/s & 1.71 m/s & \textbf{4.33} & \textbf{88.81 s} \\
        
        DIRL(-unc) & 120K & 0.72 m/s & \textbf{1.93 m/s} & 19.00 & 139.37 s & 0.62 m/s &\textbf{ 2.00 m/s} & 32.67 & 154.50 s \\
        \hline
        Human & --- & 0.99 m/s & 2.06 m/s & 0.00 & 70.71 s & 0.96 m/s & 1.97 m/s & 1.67 & 78.37 s \\
        
        \hline
        \end{tabular}
\end{table*}

\begin{figure*}[t]
    \centering
    \setlength{\abovecaptionskip}{0cm}
    \includegraphics[width = 1.8\columnwidth]{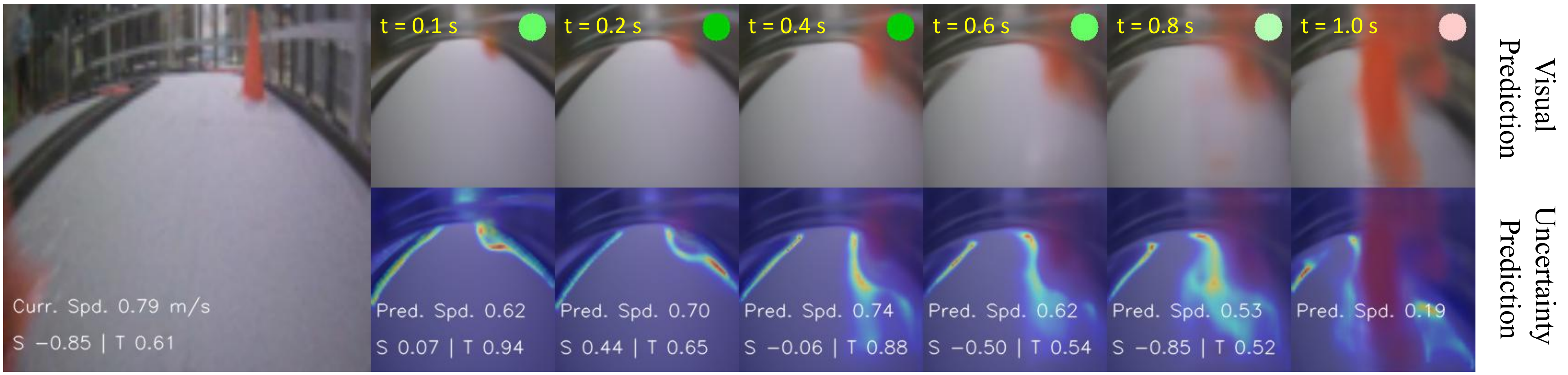}
    \caption{The agent has an imagined collision with an obstacle during policy training with \textit{Reveries-net}. During this period, the actions are first sampled from the policy network, based on which the world model predicts forward for 10 steps and returns future visual, speed and collision information with their uncertainties. The green circle indicates a safe state and the red circle means a risk state. Deeper colors of these circles mean the model is more certain (i.e., lower uncertainty) about its prediction. S and T mean steering and throttle, respectively.}
    \label{fig:reveries}
    \vspace{-0.35cm}
\end{figure*}

In this section, we evaluate the performance of our method on the real-world RC-car platform introduced in Sec. \ref{sec:system}. Our method finally takes three iterations in Algorithm \ref{alg:dirl} to achieve convergence. We also train an ablated model named DIRL(-unc) by removing the uncertainty loss in $J_t$ of Eq. (\ref{eq:policy_loss}) to measure the importance of uncertainty estimation in our method. In addition, we evaluate a human-controlled racing car to measure the performance gap of different models. 

\subsubsection{Evaluation Metrics}
We evaluate different models on two tasks with two levels of difficulty: \textit{easy} (two obstacles) and \textit{hard} (eight obstacles). The colors and positions of the obstacles are randomly chosen and fixed for each task for fair comparison. Different models are tested over three trials in a certain task, with each trial corresponding to five continuous laps of autonomous racing. During evaluation, if the agent approaches a failure mode, i.e., collisions, it will be reset by a human before continuing to run. Based on these requirements, we compute \textit{four metrics} for analysis: average speed, top speed, intervention number and time cost (reset time excluded), which are averaged over the three independent trials.

\subsubsection{Quantitative Analysis}
\label{subsub:quantitative}
The quantitative results of the different methods are shown in Table \ref{tab:real_eval}. First, it can be seen that by iteratively collecting experience and refining the model, our method can gain self-improvement without any expert control signals. For example, DIRL(3 iter) only requires 1.33 interventions in the \textit{easy} task and 4.33 interventions in the \textit{hard} task; both are lower than its previous versions. It also achieves the lowest time for 5 laps of racing. Second, by removing the uncertainty loss during policy training, DIRL(-unc) significantly degrades the performance. For example, DIRL(-unc) costs as much as 154.5 s and 32.67 interventions to finish the \textit{hard} task, ranking last among all the models. Actually, this method tends to drive very fast without performing any driving skills (e.g., slowing down the speed at sharp U-shaped corners), seeming to be over-confident in its predictions. Therefore, we can conclude that uncertainty estimation is critical for our method, performing like a regularizing term and able to improve the safety of the learned policy. Finally, we observe that even the best DIRL(3 iter) model cannot equal the driving of human racers and takes about 10 more seconds in the \textit{hard} task.

\subsubsection{Qualitative Analysis of Reveries-net}
To improve the safety and efficiency of training for physical robots, we propose to learn policies in an \textit{offline} manner by querying world predictions in \textit{Reveries-net}. For better understanding of this model, we visualize some training steps (Line 13 of Algorithm \ref{alg:dirl}) in Fig. \ref{fig:reveries}. We can see that the model is able to make a reasonable prediction based on the current observation and input action sequence. Specifically, in Fig. \ref{fig:reveries}, the agent tries to go forward by applying high throttles from 0.1 s to 0.4 s, while the opposite red obstacle also correctly changes its relative positions in the images. Finally at $t$ = 1.0 s, the model indicates that the agent collides with the obstacle, but the predicted image is rather blurry and the risk uncertainty is high. We can also observe that the image uncertainties focus on the obstacle around $t$ = 0.8 s. These results mean the model is not very confident with this prediction. We believe such an uncertainty-aware capability is important for policy learning, which has been demonstrated in Sec. \ref{subsub:quantitative}.
\begin{figure}[t]
        \centering
        \setlength{\abovecaptionskip}{0cm}
        \includegraphics[width = \columnwidth]{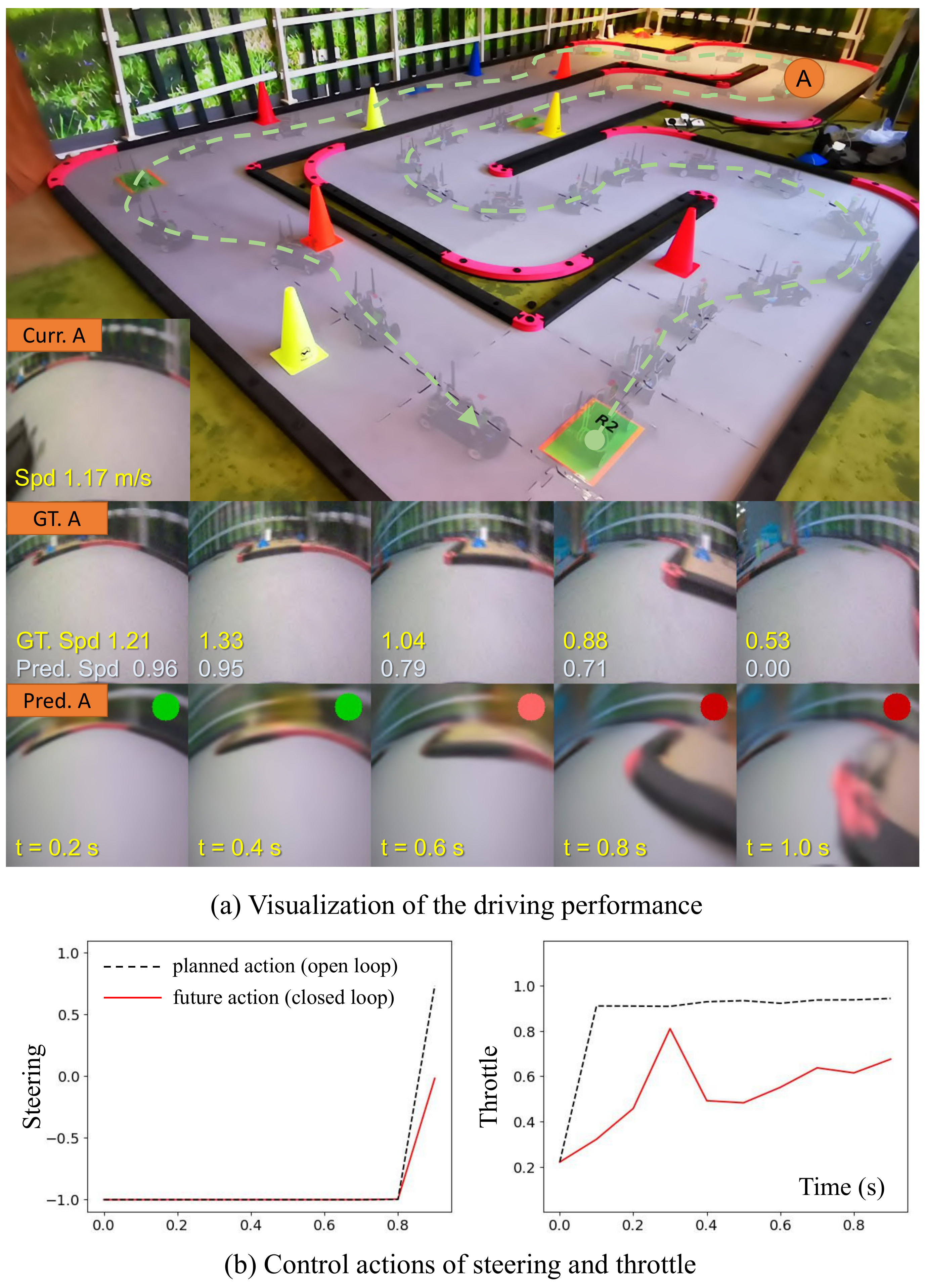}
        \caption{Driving performance of the RC-car when testing our method in the \textit{hard} task. (a) Point A is taken on the trajectory for analysis and the related reveries are shown in the bottom row with ground-truth future images for comparison. Note that the reveries are not obtained during testing but are post-rendered with current observations and open-loop actions. (b) The planned open-loop actions  (planned by the policy network at each step, but only the first one will be executed) at point A and related closed-loops actions (the actually executed actions recorded by the car).}
        \label{fig:test}
        \vspace{-0.3cm}
\end{figure}
\subsubsection{Qualitative Analysis of the Policy}
After being trained in reveries, the policy network is further evaluated in the real world. We show a lap of the RC-car trajectory in Fig. \ref{fig:test}. We can see that the agent agilely drives through multiple obstacles while staying safe on the track. We observe from the reveries of point A that the agent would collide with the fence at $t$ = 1 s in the future if it follows the open-loop actions. However, since the agent replans at every time step, it executes lower throttles than the open-loop actions in the future, and finally drives around the corner safely (GT. A in Fig. \ref{fig:test}).
\vspace{-0.1cm}

\section{Conclusion}
In this work, to achieve vision-based autonomous car racing, we presented a deep imitative reinforcement learning (DIRL) framework to train end-to-end driving policies. We unified IL and RL, where IL is used to initialize the policy, and model-based RL is used for further refinement by interacting with an uncertainty-aware world model, \textit{Reveries-net}. We quantitatively showed via simulation experiments that DIRL provides a better method of incorporating IL and RL, which is 1) 1.8$\sim$8.4 times more data efficient than previous methods, 2) beneficial for improving final task performance beyond the scope of supervised data, and 3) more robust against sub-optimal demonstrations. Finally, we evaluated DIRL on a real-world RC-car platform and performed an ablation study to show that the estimated uncertainty with evidential learning can help train much safer policies. 

In summary, our method makes a step toward bringing the RL method to real-world applications with limited data and a lower hardware burden. However, at the current stage, it still cannot match the performance of human experts. In the future, we will investigate how to achieve better driving performance by refinig the architecture of \textit{Reveries-net}. For example, the predicted visual modality can also be semantic images, which are more concise than the raw images adopted in this work.

\bibliographystyle{IEEEtran}
\bibliography{root.bib}

\end{document}